\documentclass[11pt,a4paper]{article}
\usepackage[nohyperref]{acl2017}
\usepackage{times}
\usepackage{latexsym}

\usepackage{url}
\usepackage{xspace}

\usepackage{algorithm}
\usepackage{algorithmic}

\usepackage{amsmath}
\usepackage{amssymb}

\usepackage{graphicx}
\usepackage{subcaption}
\usepackage{multirow}

\usepackage{color}
\newcommand\vvec[1]{\mathbf{#1}}

\newcommand{\lstm}{\textsc{BiLSTM}\xspace}
\newcommand{\crf}{\textsc{BiLSTM-Crf}\xspace}
\newcommand{\minitagger}{\textsc{MiniTagger}\xspace}
\newcommand{\debias}{\textsc{BiLSTM-Debias}\xspace}

\newcommand{\joint}{\textsc{Joint}\xspace}
\newcommand{\distant}{\textsc{Distant}\xspace}

\aclfinalcopy 
\def\aclpaperid{3394} 

\title{Model Transfer for Tagging Low-resource Languages using
  a Bilingual Dictionary}

\author{Meng Fang \and Trevor Cohn\\
School of Computing and Information Systems \\
The University of Melbourne\\
{\tt meng.fang@unimelb.edu.au},~{\tt t.cohn@unimelb.edu.au} \\}

\date{}

\begin{document}
\maketitle
\begin{abstract}
Cross-lingual model transfer is a compelling and popular method for
predicting annotations in a low-resource language, whereby parallel
corpora provide a bridge to a high-resource language and its
associated annotated corpora. However, parallel data is not readily
available for many languages, limiting the applicability of these
approaches. We address these drawbacks in our framework which takes
advantage of cross-lingual word embeddings trained solely on a high
coverage bilingual dictionary. We propose a novel neural network model for joint training from both sources of data based on cross-lingual word embeddings, and show substantial empirical improvements over baseline techniques. We also propose several active learning heuristics, which result in improvements over competitive benchmark methods.
\end{abstract}

\section{Introduction}
Part-of-speech (POS) tagging is an important first step in most
natural language processing (NLP) applications.
Typically this is modelled using sequence labelling methods 
to predict the conditional probability of taggings given word
sequences, using linear graphical models~\cite{lafferty2001conditional}, or  
neural network models, such as recurrent neural networks (RNN)~\cite{mikolov2010recurrent,huang2015bidirectional}.
These supervised learning algorithms rely on large labelled corpora;
this is particularly true for state-of-the-art neural network models.
Due to the expense of annotating sufficient data, such techniques are
not well suited to applications in low-resource languages.

Prior work on low-resource NLP has primarily focused on exploiting parallel corpora to
project information between a high- and low-resource language~\cite{yarowsky2001inducing,tackstrom2013token,guo2015cross,agic2016multilingual,buys2016cross}.
For example, POS tags can be projected via word alignments, and the
projected POS is then
used to train a model in the low-resource language~\cite{das2011unsupervised,zhang2016ten,fang2016learn}.
These methods overall have limited effectiveness due to errors in the
alignment and fundamental differences between the languages. 
They also assume a large parallel corpus, which may not be available for many low-resource languages.

To address these limitations, we propose a new technique for low
resource tagging, with more modest resource requirements:
1) a bilingual dictionary; 
2) monolingual corpora in the high and low resource languages;
and
3) a small annotated corpus of around $1,000$ tokens in the low-resource
language.
The first two resources are used as a form of distant supervision
through learning cross-lingual
word embeddings over the monolingual corpora and bilingual
dictionary~\cite{ammar2016massively}. 
Additionally, our model jointly incorporates the
language-dependent information from the small set of gold annotations.
Our approach combines these two sources of supervision using 
multi-task learning, such that the kinds of errors that occur in
cross-lingual transfer can be accounted for, and corrected automatically.

\begin{figure*}
\centering
\includegraphics[width=0.75\textwidth]{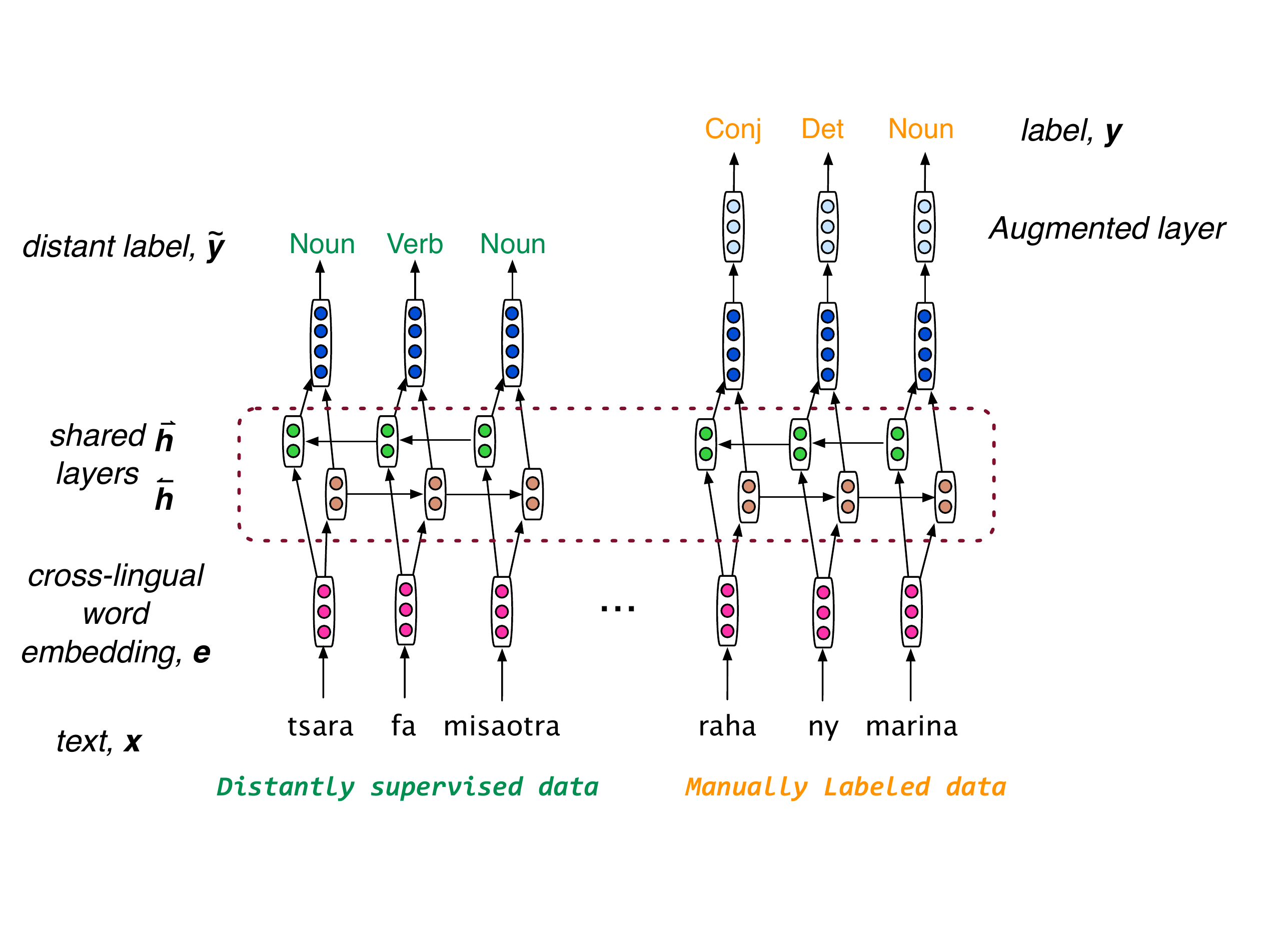}
\vspace{-6ex}
\caption{Illustration of the architecture of the joint model, which performs joint inference over both distant supervision (left) and manually labelled data (right).}
\label{fig-dist}
\end{figure*}

We empirically demonstrate the validity of our observation by using
distant supervision to improve POS tagging performance with little
supervision. Experimental results show the effectiveness of our
approach across several low-resource languages, including both
simulated and true low-resource settings. Furthermore, given the clear superiority of
training with manual annotations, we compare several active learning
heuristics. Active learning using uncertainty sampling with a
word-type bias 
leads to substantial gains
over benchmark methods such as token or sentence level uncertainty 
sampling.

\section{Related work}
POS tagging has been studied for many years. Traditionally, probabilistic models are a popular choice, such as Hidden Markov Models (HMM) and Conditional Random Fields (CRF)~\cite{lafferty2001conditional}. 
Recently, neural network models have been developed for POS tagging and achieved good performance, such as RNN and bidirectional long short-term memory (BiLSTM) and CRF-BiLSTM models~\cite{mikolov2010recurrent,huang2015bidirectional}.  
For example, the CRF-BiLSTM POS tagger obtained the state-of-the-art performance on Penn Treebank WSJ corpus~\cite{huang2015bidirectional}. 

However, in low-resource languages, these models are seldom used because of limited labelled data. Parallel
data therefore appears to be the most realistic additional source of information for developing NLP systems in low-resource languages~\cite{yarowsky2001inducing,das2011unsupervised,tackstrom2013token,fang2016learn,zhang2016ten}.
~\newcite{yarowsky2001inducing} pioneered the use of parallel data for projecting POS tag information from one language to another language. 
~\newcite{das2011unsupervised} used parallel data and exploited graph-based label propagation to expand the coverage of labelled tokens. 
~\newcite{tackstrom2013token} constructed tag dictionaries by projecting tag information from a high-resource language to a low-resource language via alignments in the parallel text.
~\newcite{fang2016learn} used parallel data to obtain projected tags as distant labels and proposed a joint BiLSTM model trained on both the distant data and $1,000$ tagged tokens. 
~\newcite{zhang2016ten} used a few word translations pairs to find a linear transformation between two language embeddings. Then they used unsupervised learning to refine embedding transformations and model parameters. Instead we use minimal supervision to refine `distant' labels through modelling the tag transformation, based on a small set of annotations.

\section{Model} \label{sec:method}

We now describe the modelling framework for POS tagging in a
low-resource language, based on very limited linguistic resources.
Our approach extends the work of~\newcite{fang2016learn}, who present a model
based on distant supervision in the form of cross-lingual projection and use projected tags generated from parallel corpora as distant annotations.
There are three main differences between their work and ours: 1) We do not
use parallel corpora, but instead use a bilingual dictionary for
knowledge transfer. 
2) Our model uses a more expressive multi-layer
perceptron when generating the gold standard tags. The multi-layer
perceptron can capture both language-specific information
and consistent tagging errors arising from this method of supervision. 
3) We propose a number of active learning
methods to further reduces the annotation requirements.
Our method is illustrated in Figure~\ref{fig-dist}, and we now
elaborate on the model components.

\paragraph{Distant cross-lingual supervision}
In order to transfer tag information between the high- and
low-resource languages, we start by learning cross-lingual word
embeddings, which operate by learning vector valued embeddings such
that words and their translations tend to be close together in the
vector space. We use the embeddings from \newcite{ammar2016massively} which trains
monolingual \texttt{word2vec} distributional representations, which are then projected into a common space, learned from bilingual dictionaries.

\begin{figure}[t]
\centering
\includegraphics[width=0.35\textwidth]{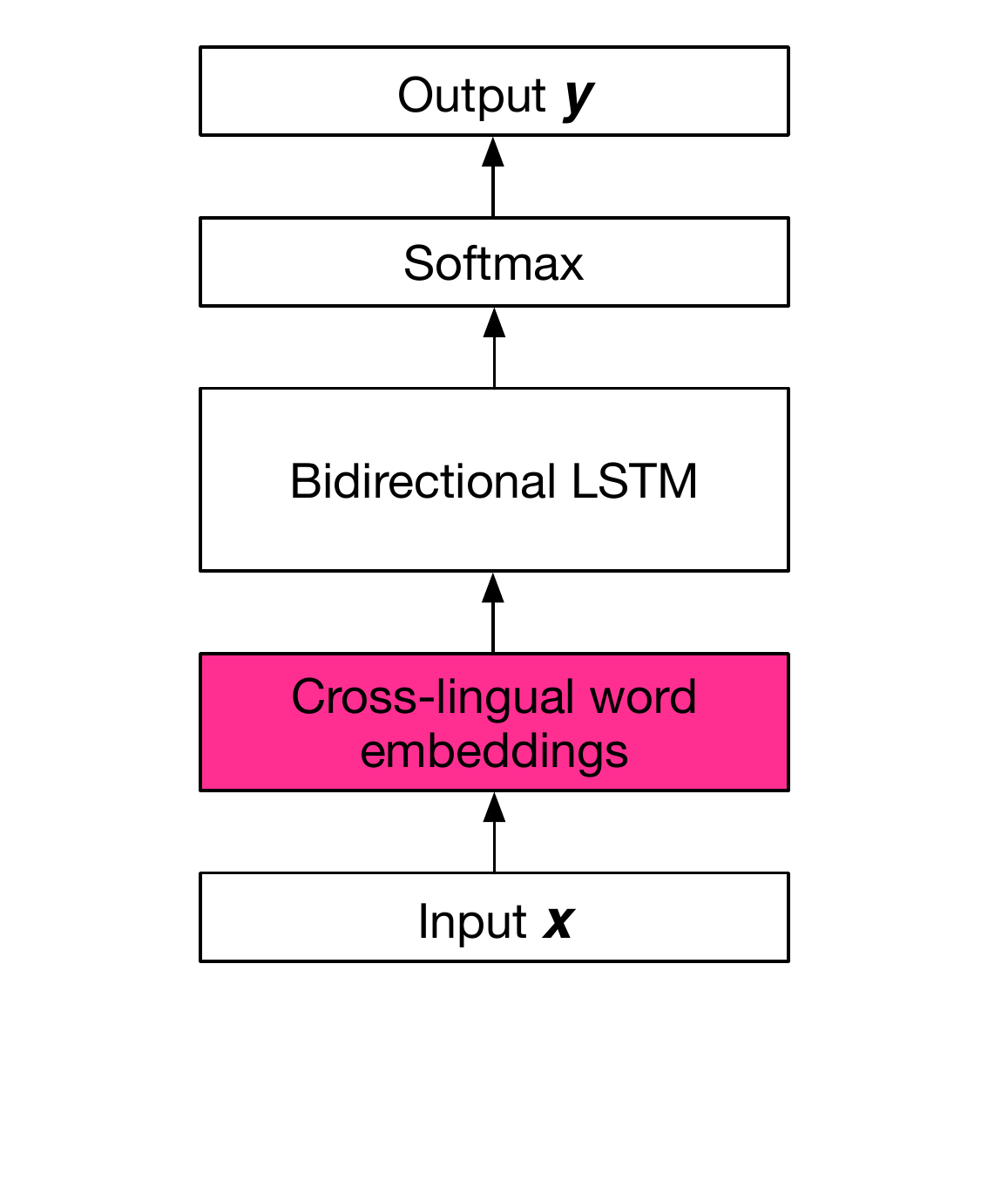}
\vspace{-7ex}
\caption{Architecture of the universal POS tagger. Cross-lingual word embeddings are pretrained using monolingual corpora and bilingual dictionaries.}
\label{fig-unitagger}
\end{figure}

We then train a POS tagger on the high-resource language, 
using the cross-lingual word embeddings as the first, fixed, layer of
a bidirectional LSTM tagger.
The tagger is a language-universal model based on cross-lingual word embeddings, for processing an arbitrary language, given a monolingual corpus and a bilingual dictionary, as shown in Figure~\ref{fig-unitagger}. Next we apply this tagger to unannotated
text in the low-resource language; this application is made possible
through the use of cross-lingual word embeddings. We refer to text tagged
this way as \emph{distantly supervised data}, and emphasize that
although much better than chance, the outputs are often incorrect and
are of limited utility on their own.

As illustrated in Figure~\ref{fig-dist}, the distant components are
generated directly as softmax outputs, $y_t \sim
\operatorname{Categorial}(\vvec{o}_t)$, with parameters 
$\vvec{o}_t = \operatorname{Softmax}(W \vvec{h}_t + \vvec{b})$
as a linear classifier over a sentence encoding, $\vvec{h}_t$, which is the output of a bidirectional LSTM encoder over the words.

\paragraph{Ground truth supervision}
The second component of the model is manually labelled
text in the low-resource language. To model this data
we employ the same model structure as above but augmented with a second perceptron output
layer, as illustrated in Figure~\ref{fig-dist} (right).
Formally, $\tilde{y}_t \sim \operatorname{Categorial}(\tilde{\vvec{o}}_t)$ where 
$\tilde{\vvec{o}}_t = \operatorname{MLP}(\vvec{o}_t)$ is a single hidden layer
perceptron with $\tanh$ activation and softmax output transformation.
This component allows for a more expressive label mapping than 
\newcite{fang2016learn}'s linear matrix translation.

\paragraph{Joint multi-task learning}
To combine the two sources of information, we use a joint 
objective, 
\begin{equation} 
\mathcal{J}= 
-
\gamma \sum_{t \in \mathcal{N}} \langle
\tilde{y}_{t} ,
\log  \tilde{o}_t \rangle -  
\sum_{t \in \mathcal{M}} \langle 
y_{t} ,
\log o_t \rangle
\, ,
\end{equation}
where $\cal N$ and $\cal M$ index the token positions in the distant and ground
truth corpora, respectively, and $\gamma$ is a constant balancing the
two components which we set for uniform weighting,
$\gamma = \cal \frac{|M|}{|N|}$.

Consider the training effect of the true POS tags: when performing
error backpropagation, the cross-entropy error signal must pass
through the transformation linking $\tilde{o}$ with $o$, which can be
seen as a language-specific step, after which the generalised error signal can be further backpropagated to the rest of the model.

\paragraph{Active learning}
Given the scarcity of ground truth labels and the high cost of annotation, a natural question is whether we can optimise which text to be annotated in order achieve the high accuracy for the lowest cost. 
We now outline a range of active learning approaches based on the following heuristics, which are used to select the instances for annotation from a pool of candidates:
\begin{description}
\item[\textsc{Token}] Select the token $x_t$  with the highest uncertainty, \mbox{$H(\vvec{x},t) = -\sum_{y} P(y|\vvec{x}, t) \log P(y|\vvec{x}, t)$};
\item[\textsc{Sent}] Select the sentence $\vvec{x}$  with the highest aggregate uncertainty, $H(\vvec{x}) = \sum_{t} H(\vvec{x}, t)$;
\item[\textsc{FreqType}] Select the most frequent unannotated word  type~\cite{garrette2013learning}, in which case all token instances are annotated with the most frequent label for the type in the training corpus;\footnote{We could support more than one class label, by marginalising over the set of valid labels for all tokens in the training objective.} 
\item[\textsc{SumType}] Select a word type, $z$, for annotation with the highest aggregate uncertainty over token occurrences, \mbox{$H(z) = \sum_{i \in \mathcal{D}} \sum_{x_{i,t}=z} H(\vvec{x}_i, t)$}, which effectively combines uncertainty sampling with a bias towards high frequency types; and
\item[\textsc{Random}] Select word types randomly.
\end{description}

\begin{table*}
\footnotesize
\begin{center}
\begin{tabular}{l c c c c c c c  c | c c}
\hline ~  & da & nl & de & el  & it & pt & es & sv & tk & mg   \\ \hline
Random & 23.2 & 30.5   & 27.1  & 23.2& 25.9  & 24.3& 26.9 & 21.6 & 36.9 & 34.5   \\
\lstm & 61.8 & 62.1   & 60.5 & 70.1 & 73.6   & 67.6 & 63.6& 57.2 & 44.0 & 63.4 \\
\crf & 46.3 & 47.7  & 53.2 & 35.1 & 41.2   & 44.1 & 25.5& 54.9  & 43.1 & 41.4   \\
\minitagger & 77.0 & 72.5 & 75.9  & 75.7 & 67.3   & 75.1 & 73.5 & 77.7  &  49.8 & 67.2  \\ 
 \hline

\distant\scriptsize{+CCA} &  73.5 & 64.5 & 57.7  & 53.1 & 59.5   & 67.8 & 63.5 &  66.0 & 57.2 & 49.7   \\
\distant\scriptsize{+Cluster}  & 70.4 &  61.7 &  65.9  &  65.5 &  64.8  &  66.9&  68.4  & 64.1 & 51.7 & 50.2   \\
\hline
\debias\scriptsize{+CCA} & 73.2 & 72.8 & 72.5  & 71.2 & 70.7  & 72.1  & 71.1 & 73.1  &  49.2 & 65.9  \\
\debias\scriptsize{+Cluster} & 72.5 & 70.1 & 71.2  & 68.7 & 69.1  & 72.5  & 70.6 & 73.3  &  48.7 & 64.5  \\
\joint\scriptsize{+CCA}  & 81.1  & \bf 82.3 & 76.1  & 77.5 & 75.9  &\bf 82.1 & 79.7  & \bf 78.1 &\bf 72.6 & 75.3  \\
\joint\scriptsize{+Cluster} & \bf 81.9 & 81.5   & \bf 78.9  & \bf 80.1 & \bf 81.9   & 76.7 & \bf 81.2 & 78.0 & 70.4 & \bf 75.7   \\ 
\hline
\end{tabular}
\end{center}
\vspace{-1ex}
\caption{POS tagging accuracy on over the ten target languages, showing first approaches using only the gold data; next methods using only distant cross-lingual supervision, and lastly joint multi-task learning. English is used as the source language and columns correspond to a specific target language. }
\label{tab-uni}
\end{table*}

\section{Experiments}
We evaluate the effectiveness of the proposed model for several different languages, including both simulated low-resource and true low-resource settings. The first evaluation set uses the CoNLL-X datasets of European languages \cite{buchholz2006conll}, comprising Danish (da), Dutch (nl), German (de), Greek (el), Italian (it), Portuguese (pt), Spanish (es) and Swedish (sv). We use the standard corpus splits. The first 20 sentences of training set are used for training as the tiny labelled (gold) data and the last 20 sentences are used for development (early stopping). 
We report accuracy on the held-out test set.

The second evaluation set includes two highly challenging languages, Turkish (tk) and Malagasy (mg), both having high morphological complexity and the latter has truly scant resources.
Turkish data was drawn from CoNLL
2003\footnote{http://www.cnts.ua.ac.be/conll2003/ner/} and Malagasy
data was collected from \newcite{das2011unsupervised}, in both cases using the same training configuration as above. 

In all cases English is used as the source `high resource' language, on which we train a tagger using the Penn Treebank, and we evaluate on each of the remaining languages as an independent target. For cross-lingual word embeddings, we evaluate two techniques from~\newcite{ammar2016massively}: CCA-based word embeddings and cluster-based word embeddings. Both types of word embedding techniques are based on bilingual dictionaries. The dictionaries were formed by translating the $20k$ most common words in the English monolingual corpus with Google Translate.\footnote{Although the use of a translation system conveys a dependence on parallel text, high quality word embeddings can be learned directly from bilingual dictionaries such as Panlex \cite{kamholz2014panlex}.} The monolingual corpora were constructed from a combination of text from the Leipzig Corpora Collection and Europarl. We trained the language-universal POS tagger based on the cross-lingual word embeddings with the universal POS tagset \cite{petrov2011universal}, and then applied to the target language using the embedding lookup table for the corresponding language embeddings. We implement our learning procedure with the DyNet toolkit~\cite{dynet}.\footnote{Code available at \url{https://github.com/mengf1/trpos}} The BiLSTM layer uses $128$ hidden units, and  $32$ hidden units for the transformation step. We used SGD with momentum to train models, with early stopping based on development performance. 

For benchmarks, we compare the proposed model against various
state-of-the-art supervised learning methods, namely: a \lstm tagger,
\crf tagger~\cite{huang2015bidirectional}, and a state-of-the-art semi-supervised POS tagging algorithm,
\minitagger~\cite{stratos2015simple}, which is also focusing on
minimising the amount of labelled data. Note these methods do not use
cross-lingual supervision. For a more direct comparison, we include \debias~\cite{fang2016learn},
applied using our proposed cross-lingual supervision based on dictionaries, instead of parallel corpora; accordingly the key difference is their linear transformation for the distant data, versus our non-linear transformation to the gold data.

\begin{figure*}[t]
\centering
\includegraphics[width=0.75\textwidth]{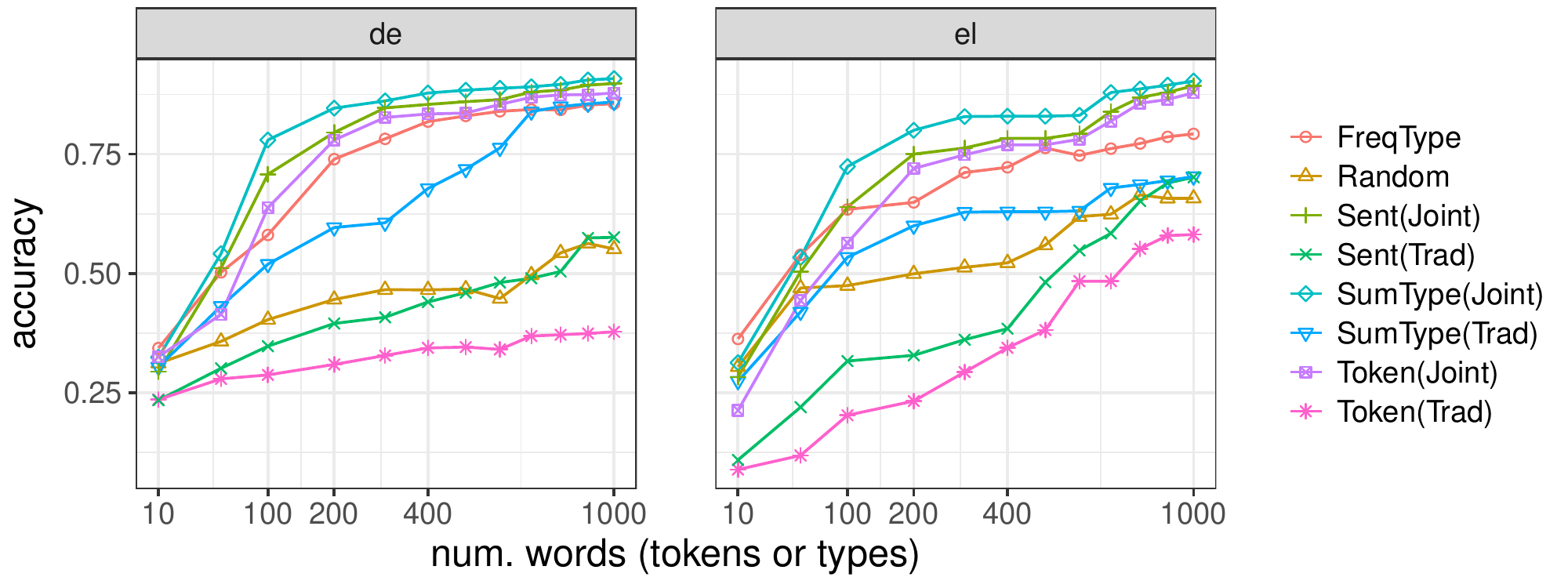}
\vspace{-1ex}
\caption{Active learning evaluation on German and Greek, using CCA
trained cross-lingual word embeddings. Trad means traditional
active learning; Joint means joint multi-task learning.}
\label{fig-al}
\end{figure*}

\paragraph{Results}

Table~\ref{tab-uni}  reports the tagging accuracy, showing that our models consistently outperform the baseline techniques. 
The poor performance of the supervised methods suggests they are overfitting the small training set, however this is much less of a problem for our approach (labelled Joint).
Note that distant supervision alone gives reasonable performance
(labelled \distant) however the joint modelling of the ground truth and
distant data yields significant improvements in almost all
cases. \debias~\cite{fang2016learn} performs worse than
our proposed method, indicating that a linear transformation is
insufficient for modelling distant supervision.
The accuracies are higher overall for the European cf. Turkic
languages, presumably because these languages are closer to English,
have higher quality dictionaries and in most cases are morphologically
simpler. 
Finally, note the difference between CCA and Cluster methods for learning word embeddings which arise from the differing quality of distant supervision between the languages.

Figure~\ref{fig-al} compares various active learning heuristics (see \S\ref{sec:method})
based on different taggers, either a supervised \lstm (labelled Trad) or our
multi-task model which also includes cross-lingual supervision
(\joint). 

Traditional uncertainty-based sampling strategies
(\textsc{Token}(Trad) and \textsc{Sent}(Trad)) do not work well
because models based on limited supervision do not provide accurate
uncertainty information,\footnote{Sentence level annotation is likely to be much faster  than token or type level annotation, however even if it were an order of magnitude faster it is still not a competitive active learning strategy.} and moreover, annotating at the type rather
than token level provides a significantly stronger supervision
signal. The difference is apparent from the decent performance of Random sampling
over word types.
Overall, \textsc{SumType}(Joint) outperforms the other heuristics
consistently, underlining the importance of cross-lingual distant
supervision, as well as combining the benefits of uncertainty
sampling, type selection and a frequency bias. 
Comparing the amount of annotation required between the best
traditional active learning method \textsc{SumType}(Trad) and our best
method \textsc{SumType}(Joint), we achieve the same performance with
an order of magnitude less annotated data ($100$~vs.~$1,000$ labelled words).

\section{Conclusion}
In this paper, we proposed a means of tagging a
low-resource language without the need for bilingual parallel
corpora. We introduced a new cross-lingual distant supervision
method based on a bilingual dictionary.
Furthermore, deep
neural network models can be effective with limited supervision by
incorporating distant supervision, in the form of model transfer with
cross-lingual word embeddings. We show that traditional uncertainty
sampling strategies do not work well on low-resource settings, and
introduce new methods based around labelling word types. 
Overall our approach leads to consistent and substantial improvements over benchmark
methods. 

\section*{Acknowledgments}
This work was sponsored by the Defense Advanced Research Projects Agency Information Innovation Office (I2O) under the Low Resource Languages for Emergent Incidents (LORELEI) program issued by DARPA/I2O under Contract No. HR0011-15-C-0114. The views expressed are those of the author and do not reflect the official policy or position of the Department of Defense or the U.S. Government. Trevor Cohn was supported by the Australian Research Council Future Fellowship (project number FT130101105).

\bibliography{acl2017}

\begin{thebibliography}{}
\expandafter\ifx\csname natexlab\endcsname\relax\def\natexlab#1{#1}\fi

\bibitem[{Agi{\'c} et~al.(2016)Agi{\'c}, Johannsen, Plank, Mart{\'\i}nez,
  Schluter, and S{\o}gaard}]{agic2016multilingual}
{\v{Z}}eljko Agi{\'c}, Anders Johannsen, Barbara Plank, H{\'e}ctor~Alonso
  Mart{\'\i}nez, Natalie Schluter, and Anders S{\o}gaard. 2016.
\newblock Multilingual projection for parsing truly low-resource languages.
\newblock {\em Transactions of the Association for Computational Linguistics\/}
  4:301--312.

\bibitem[{Ammar et~al.(2016)Ammar, Mulcaire, Tsvetkov, Lample, Dyer, and
  Smith}]{ammar2016massively}
Waleed Ammar, George Mulcaire, Yulia Tsvetkov, Guillaume Lample, Chris Dyer,
  and Noah~A Smith. 2016.
\newblock Massively multilingual word embeddings.
\newblock {\em Transactions of the Association for Computational Linguistics\/}
  4:431--444.

\bibitem[{Buchholz and Marsi(2006)}]{buchholz2006conll}
Sabine Buchholz and Erwin Marsi. 2006.
\newblock Conll-x shared task on multilingual dependency parsing.
\newblock In {\em Proceedings of the Tenth Conference on Computational Natural
  Language Learning\/}. Association for Computational Linguistics, pages
  149--164.

\bibitem[{Buys and Botha(2016)}]{buys2016cross}
Jan Buys and Jan~A. Botha. 2016.
\newblock Cross-lingual morphological tagging for low-resource languages.
\newblock In {\em Proceedings of the 54th Annual Meeting of the Association for
  Computational Linguistics (ACL)\/}. Association for Computational
  Linguistics, Berlin, Germany, pages 1954--1964.

\bibitem[{Das and Petrov(2011)}]{das2011unsupervised}
Dipanjan Das and Slav Petrov. 2011.
\newblock Unsupervised part-of-speech tagging with bilingual graph-based
  projections.
\newblock In {\em Proceedings of the 49th Annual Meeting of the Association for
  Computational Linguistics: Human Language Technologies (ACL-HLT)\/}. pages
  600--609.

\bibitem[{Fang and Cohn(2016)}]{fang2016learn}
Meng Fang and Trevor Cohn. 2016.
\newblock Learning when to trust distant supervision: An application to
  low-resource pos tagging using cross-lingual projection.
\newblock In {\em Proceedings of the 20th SIGNLL Conference on Computational
  Natural Language Learning (CoNLL)\/}. Berlin, Germany.

\bibitem[{Garrette and Baldridge(2013)}]{garrette2013learning}
Dan Garrette and Jason Baldridge. 2013.
\newblock Learning a part-of-speech tagger from two hours of annotation.
\newblock In {\em Proceedings of the 2013 Conference of the North American
  Chapter of the Association for Computational Linguistics: Human Language
  Technologies (NAACL-HLT)\/}. Citeseer, pages 138--147.

\bibitem[{Guo et~al.(2015)Guo, Che, Yarowsky, Wang, and Liu}]{guo2015cross}
Jiang Guo, Wanxiang Che, David Yarowsky, Haifeng Wang, and Ting Liu. 2015.
\newblock Cross-lingual dependency parsing based on distributed
  representations.
\newblock In {\em Proceedings of the 53rd Annual Meeting of the Association for
  Computational Linguistics (ACL)\/}. Association for Computational
  Linguistics, pages 1234--1244.

\bibitem[{Huang et~al.(2015)Huang, Xu, and Yu}]{huang2015bidirectional}
Zhiheng Huang, Wei Xu, and Kai Yu. 2015.
\newblock Bidirectional lstm-crf models for sequence tagging.
\newblock {\em arXiv preprint arXiv:1508.01991\/} .

\bibitem[{Kamholz et~al.(2014)Kamholz, Pool, and Colowick}]{kamholz2014panlex}
David Kamholz, Jonathan Pool, and Susan~M Colowick. 2014.
\newblock Panlex: Building a resource for panlingual lexical translation.
\newblock In {\em Proceedings of the Ninth International Conference on Language
  Resources and Evaluation (LREC)\/}. pages 3145--3150.

\bibitem[{Lafferty et~al.(2001)Lafferty, McCallum, and
  Pereira}]{lafferty2001conditional}
John Lafferty, Andrew McCallum, and Fernando Pereira. 2001.
\newblock Conditional random fields: Probabilistic models for segmenting and
  labeling sequence data.
\newblock In {\em Proceedings of the 8th International Conference on Machine
  Learning (ICML)\/}. volume~1, pages 282--289.

\bibitem[{Mikolov et~al.(2010)Mikolov, Karafi{\'a}t, Burget, Cernock{\`y}, and
  Khudanpur}]{mikolov2010recurrent}
Tomas Mikolov, Martin Karafi{\'a}t, Lukas Burget, Jan Cernock{\`y}, and Sanjeev
  Khudanpur. 2010.
\newblock Recurrent neural network based language model.
\newblock In {\em Interspeech\/}. volume~2, page~3.

\bibitem[{Neubig et~al.(2017)Neubig, Dyer, Goldberg, Matthews, Ammar,
  Anastasopoulos, Ballesteros, Chiang, Clothiaux, Cohn, Duh, Faruqui, Gan,
  Garrette, Ji, Kong, Kuncoro, Kumar, Malaviya, Michel, Oda, Richardson,
  Saphra, Swayamdipta, and Yin}]{dynet}
Graham Neubig, Chris Dyer, Yoav Goldberg, Austin Matthews, Waleed Ammar,
  Antonios Anastasopoulos, Miguel Ballesteros, David Chiang, Daniel Clothiaux,
  Trevor Cohn, Kevin Duh, Manaal Faruqui, Cynthia Gan, Dan Garrette, Yangfeng
  Ji, Lingpeng Kong, Adhiguna Kuncoro, Gaurav Kumar, Chaitanya Malaviya, Paul
  Michel, Yusuke Oda, Matthew Richardson, Naomi Saphra, Swabha Swayamdipta, and
  Pengcheng Yin. 2017.
\newblock Dynet: The dynamic neural network toolkit.
\newblock {\em arXiv preprint arXiv:1701.03980\/} .

\bibitem[{Petrov et~al.(2011)Petrov, Das, and McDonald}]{petrov2011universal}
Slav Petrov, Dipanjan Das, and Ryan McDonald. 2011.
\newblock A universal part-of-speech tagset.
\newblock {\em arXiv preprint arXiv:1104.2086\/} .

\bibitem[{Stratos and Collins(2015)}]{stratos2015simple}
Karl Stratos and Michael Collins. 2015.
\newblock Simple semi-supervised pos tagging.
\newblock In {\em Proceedings of the 2015 Conference of the North American
  Chapter of the Association for Computational Linguistics: Human Language
  Technologies (NAACL-HLT)\/}. pages 79--87.

\bibitem[{T{\"a}ckstr{\"o}m et~al.(2013)T{\"a}ckstr{\"o}m, Das, Petrov,
  McDonald, and Nivre}]{tackstrom2013token}
Oscar T{\"a}ckstr{\"o}m, Dipanjan Das, Slav Petrov, Ryan McDonald, and Joakim
  Nivre. 2013.
\newblock Token and type constraints for cross-lingual part-of-speech tagging.
\newblock {\em Transactions of the Association for Computational Linguistics\/}
  1:1--12.

\bibitem[{Yarowsky and Ngai(2001)}]{yarowsky2001inducing}
David Yarowsky and Grace Ngai. 2001.
\newblock Inducing multilingual {POS} taggers and {NP} brackets via robust
  projection across aligned corpora.
\newblock In {\em Proceedings of the 2001 Conference of the North American
  Chapter of the Association for Computational Linguistics (NAACL)\/}.

\bibitem[{Zhang et~al.(2016)Zhang, Gaddy, Barzilay, and
  Jaakkola}]{zhang2016ten}
Yuan Zhang, David Gaddy, Regina Barzilay, and Tommi Jaakkola. 2016.
\newblock Ten pairs to tag--multilingual pos tagging via coarse mapping between
  embeddings.
\newblock In {\em Proceedings of the 2016 Conference of the North American
  Chapter of the Association for Computational Linguistics: Human Language
  Technologies (NAACL-HLT)\/}. pages 1307--1317.

\end{thebibliography}
\bibliographystyle{acl_natbib}

\end{document}